\documentclass{svproc}
\usepackage{url}
\usepackage{graphicx}
\usepackage{amsmath}
\usepackage{amssymb}
\usepackage{mathrsfs}
\usepackage{optidef}
\usepackage{xcolor}
\usepackage[colorlinks = true,
            linkcolor = blue,
            urlcolor  = magenta,
            citecolor = blue,
            anchorcolor = pink]{hyperref}

\begin{document}
\mainmatter              
\title{The Surprising Effectiveness of Linear Models for Visual Foresight in Object Pile Manipulation}
\titlerunning{Effectiveness of Linear Models for Visual Foresight in Pile Manipulation}  
%
\author{H.J. Terry Suh\inst{1} \and Russ Tedrake\inst{1} \footnote{This work was supported by NSF Award No. EFMA-1830901}}
\authorrunning{H.J. Terry Suh and Russ Tedrake} 
%
\tocauthor{H.J. Terry Suh, Russ Tedrake}
\institute{Massachusetts Institute of Technology, Cambridge, MA 02139, USA,\\
\email{hjsuh@mit.edu}, \email{russt@mit.edu}}

\maketitle              

\begin{abstract}
In this paper, we tackle the problem of pushing piles of small objects into a desired target set using visual feedback. Unlike conventional single-object manipulation pipelines, which estimate the state of the system parametrized by pose, the underlying physical state of this system is difficult to observe from images. Thus, we take the approach of reasoning directly in the space of images, and acquire the dynamics of visual measurements in order to synthesize a visual-feedback policy. We present a simple controller using an image-space Lyapunov function, and evaluate the closed-loop performance using three different class of models for image prediction: deep-learning-based models for image-to-image translation, an object-centric model obtained from treating each pixel as a particle, and a switched-linear system where an action-dependent linear map is used. Through results in simulation and experiment, we show that for this task, a linear model works surprisingly well -- achieving better prediction error, downstream task performance, and generalization to new environments than the deep models we trained on the same amount of data. We believe these results provide an interesting example in the spectrum of models that are most useful for vision-based feedback in manipulation, considering both the quality of visual prediction, as well as compatibility with rigorous methods for control design and analysis. Project site: \href{https://sites.google.com/view/linear-visual-foresight/home}{https://sites.google.com/view/linear-visual-foresight/home}

\end{abstract}
\section{Introduction}
The ability to predict the future is paramount to design and verification of a control policy, as modeling the dynamics of a system can enable planning and control approaches that solve tasks using the same model, or can facilitate analyzing the behavior of the closed-loop system. Conventionally in robotics, such dynamics are obtained using the laws of physics, and the state of the system is often defined as generalized coordinates of Lagrangian dynamics \cite{mason1}. 

However, it is not always clear how to estimate the physical state, or the dynamics of how those states evolve, directly from sensor measurements. A planar pushing task \cite{mason1} is a great canonical example: not only is the center of mass of the object hard to observe from vision without the assumption of uniform density \cite{Tsujio1}, but the dynamics of the object is also unknown due to the uncertain interaction between the object and the table \cite{rodriguez2}. On the other hand, directly identifying the dynamics of measurements (output dynamics) offer a very general approach to vision-based control, as it relieves the need of defining true states of a system, or carefully designing an observer from visual input \cite{Finn1}. 

\begin{figure}[t]
\centering
\includegraphics[width=0.7\linewidth]{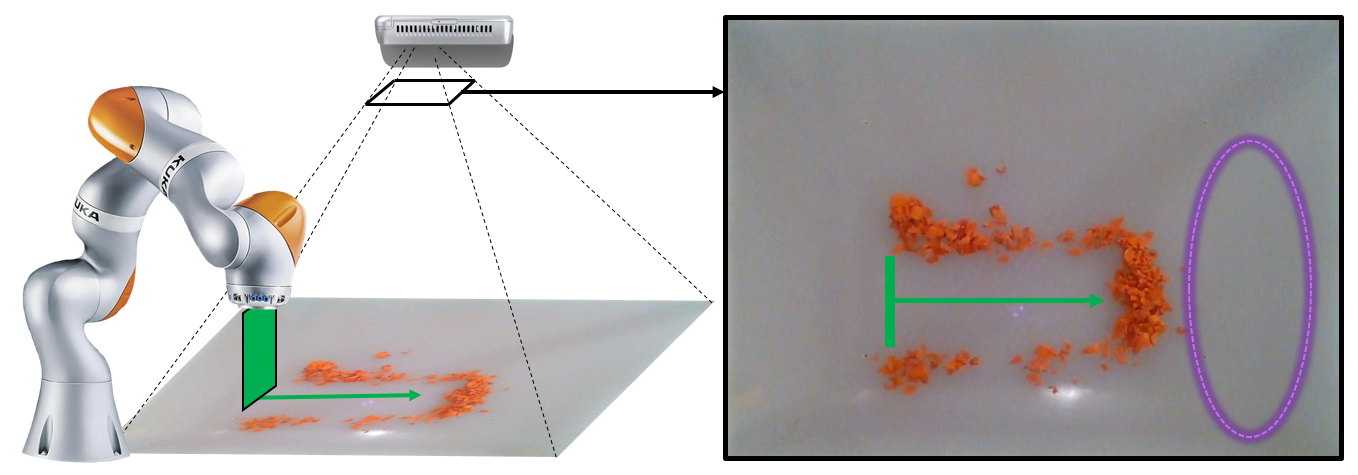}
\caption{Description of the task, where the robot must push all the carrots to fit in the purple target set. This real image illustrates the actual complexity of the phenomena, as well as the difficulty of observing states of each piece of carrot from the image.}
\label{fig:problemsetup}
\end{figure}

In this work, we deal with a task which epitomizes the strength of designing policies directly in the space of measurements: a robot observes a cutting board with diced carrots, and must find a sequence of push actions to collect them into a desired target set. The underlying physical state of the system, and even the cardinality of the state-space, is very difficult to observe from vision (Fig. \ref{fig:problemsetup}). Despite the difficulty of modeling the problem in state space, humans trivially move around piles with ease in cooking, where diced onions and carrots are moved to make room for other ingredients. We hypothesize that this is a case of output feedback, where dynamic identification and control happens directly in the space of visual measurements.

While direct output feedback offers many advantages, the dynamics of measurements must often be obtained in a data-driven manner, as it bypasses observation of physical states. Combined with the success of deep learning in vision  \cite{cnn}, recent works in vision-based control have heavily utilized deep-learning approaches. In \cite{Finn1}, the term ``Visual Foresight'' is first used to describe output dynamics of vision, and an end-to-end neural network architecture for image-to-image translation is presented. Other works \cite{prom,dvf2} also treat the output dynamics problem as an instance of image-to-image translation, and set up a deep-learning architecture to predict future frames.

Due to the difficulty of identifying output dynamics in high-dimensional pixel space, many of the recent works in vision-based control and intuitive physics have argued that instead of identifying the output dynamics, a neural network-based observer can observe the object-centric state of the system \cite{spatialautoencoder,billiard,gupta1,vin}. Since object-centric states are lower dimensional and have more identifiable continuous dynamics in their tasks, they were successful in showing that this approach leads to better performance and sample efficiency. However, we challenge the generality of this approach since object states are not always of lower dimension or have simpler dynamics compared to visual dynamics. This is exemplified by our task of pushing piles of small objects, or manipulating deformable objects such as clothes or fluids. Similarly, other works have focused on identifying dynamics over keypoints \cite{keypoints,keto}, but it is not clear how to generalize keypoints to this problem where there are multiple small objects in a pile. 

In \cite{Tucker}, an object-centric approach using graph neural networks \cite{vin} is combined with visual feature vectors to manipulate piles of objects to a desired target set. While this approach may work if there are countable number of objects in the scene, they may fail to deal with partial observation of the scene due to occlusion, or a case where million particles of sand or water must be manipulated. In \cite{8460915}, a robot finds feasible plans to manipulate piles of dirt into a target region using A$^*$ search, with a learned Random Forest transition model over a grid-like representation. However, we have questioned if this is really the right model to fit output dynamics, and focused our attention to which class of models are more adequate for capturing the dynamics of measurements in pixel-space.

We had started to explore this problem with a deep-learning-based approach, where the dynamics are estimated in the latent space of visual feature vectors. While the prediction was visually plausible, we found that we could not succeed in achieving the downstream task due to small (in the sense of mean-squared), yet critical errors that do not agree with physical behavior. Then, we wondered how well a simple linear model would work. A linear model would offer principled ways of estimating dynamics, and provide better connection with rigorous approaches in linear systems theory. The promise of Koopman operators \cite{williams_datadriven_2015} and occupation measures \cite{lasserre_nonlinear_2008} is that all dynamics become linear in high enough dimensions; perhaps the pixel coordinates are playing a similar role.

To investigate our question, we set up a simple controller using a Lyapunov function that operates directly on images, and evaluate the closed-loop behavior of three models: a switched-linear model where the linear map from image to image is a function of discrete inputs, variations of deep-learning image-prediction models inspired from \cite{Finn1,dvf2,prom}, and an object-centric transport model that treats each pixel as a particle. Through evaluations in simulation and experiment, we find that the switched-linear model provides the best performance in image prediction, downstream task, and generalization to new environments.

\section{Problem Statement}
\subsection{Setup and Notation}
As illustrated in Fig. \ref{fig:problemsetup}, our goal is to push all carrot pieces into a desired target set using visual feedback. We assume color thresholding or background subtraction so that we have a greyscale image. We denote the original greyscale image at time $k$ as $\mathbf{I}_k\in\mathbb{R}^{N\times N}$, and its vectorized form as $y_k\in\mathbb{R}^{N^2}$. Finally, the input to the system is $u\in\mathbb{R}^4$, which is consisted of the start coordinate of the push $p_i\in \mathbb{R}^2$, the orientation of the push $\theta$, and the push length $l$ \cite{Tucker,pulkit1}. Here we assume that the push surface is always perpendicular to the push direction. Our task is to learn the discrete-time dynamics of image prediction, which can be modeled by one of the two equivalent functions in \eqref{eq:image_dynamics}.
\begin{equation}
    \mathbf{I}_{k+1}=\bar{f}(\mathbf{I}_k,u) \qquad y_{k+1}=f(y_k,u).
    \label{eq:image_dynamics}
\end{equation}

Writing down the dynamics in this form also encodes our assumption of quasi-static dynamics, where friction dominates the true dynamics of the system. Our experiments with pushing piles of chopped carrots justify this assumption. 
\subsection{A Simple Lyapunov-based Controller}
The original problem can be modeled by having continuous states $\mathbf{I}_k$ and continuous inputs $u$, where the goal is to drive $\mathbf{I}_0$ to some desired set of allowable images $\mathcal{S}_\mathbf{I}$. However, we choose to discretize the inputs by a grid in the action-space, performing direct search. This is due to the high non-convexity of planar pushing; most actions will not make contact with objects and produce zero gradients. This point is illustrated well in the billiard example of \cite{difftaichi}. This setup also allows fairness for the prediction methods by allowing them to work over the same set of inputs. Solving optimal control problems for such systems with continuous states and discrete inputs often requires combinatorial search.

Fortunately, this problem admits a simple greedy strategy. To motivate this controller, imagine that the true state of the system is known, which is consisted of each particle's position $p_i\in\mathbb{R}^2$. Let $\mathcal{X}=\{p_i\}$ denote the set of these particles. Then, let $\mathcal{S}_d$ be the desired target set, which is a subset of $\mathbb{R}^2$. Our objective is to push all the particles inside $\mathcal{S}_d$. The distance from each particle to the target set is defined by looking for the closest point in the set,
\begin{equation}
    d(p_i, \mathcal{S}_d)=\min_{p_j\in\mathcal{S}_d} \|p_i - p_j\|_p,
\end{equation}
where $\|\cdot\|_p$ is the $p$-norm, a hyperparameter in the controller. Then, we simply average all the distances, and define this as a Lyapunov function on the set.
\begin{equation}
    V(\mathcal{X})=\frac{1}{|\mathcal{X}|}\sum_{p_i\in\mathcal{X}} d(p_i, \mathcal{S}_d).
    \label{eq:lyapunovparticle}
\end{equation}
This can be interpreted as the average Chamfer distance \cite{chamfer} between a discrete set of points and a continuous target set. As a sum of norms, this function is always strictly positive everywhere except for when $\forall i , d(p_i,\mathcal{S}_d)=0$, which would mean that all the particles are inside the target set. 

To extend this Lyapunov function to images, we initialize a pre-computed distance matrix $\mathbf{D}\in\mathbb{R}^{N\times N}$ where each element represents the distance between center of pixel and the target set. Let $d$ denote the vectorized form of $\mathbf{D}$. Then, the Lyapunov function on images is evaluated as a weighted average of distances, where the weight is provided from pixels. We give two expressions for the Lyapunov function in \eqref{eq:lyapunov}, where $\odot$ represents the element-wise (Hadamard) product. 

\begin{figure}[t]
\centering
\includegraphics[height=3.0cm]{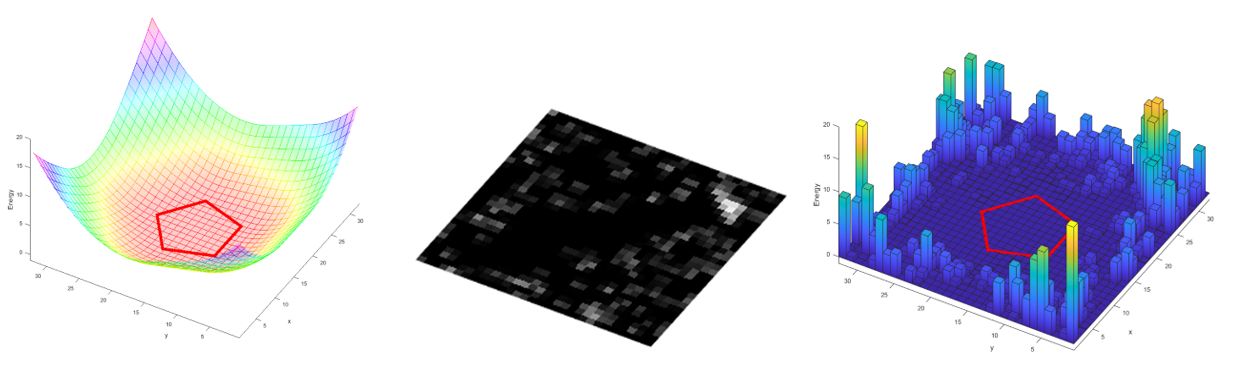}
\caption{Visualization of the Lyapunov function with a pentagonal target set and using the $p=2$ distance norm. The pre-initialized $\mathbf{D}$ matrix (left) gets multiplied with the image (center), and the average value of elements is taken from the result (right).}
\label{fig:lyapunov}
\end{figure}

\begin{equation}
    \bar{V}(\mathbf{I}_k)=\frac{1}{\|\mathbf{I}_k\|_{1,1}} \mathbf{D} \odot \mathbf{I}_k \qquad V(y_k)=\frac{1}{\|y_k\|_1}d^T y_k.
    \label{eq:lyapunov}
\end{equation}
$\bar{V}(\mathbf{I}_k)$ is zero if and only if indices of the non-zero pixels of the images coincide with the indices of zero elements of the $\mathbf{D}$ matrix, which would signify that non-zero pixels are only placed in the target set.

Finally, we argue that the $\bar{V}(\mathbf{I}_k)$ is a Control Lyapunov Function (CLF), since it satisfies the following property, if we assume $\bar{f}$ is known: 
\begin{equation}
    \forall \mathbf{I}\notin \mathcal{S}_\mathbf{I}, \quad \exists u \quad \text{s.t.} \quad \Delta \bar{V} = \bar{V}(\mathbf{I}_{k+1})-\bar{V}(\mathbf{I}_k)=\bar{V}(\bar{f}(\mathbf{I}_k,u)) - \bar{V}(\mathbf{I}_k) < 0,
\end{equation}
where $\mathcal{S}_\mathbf{I}$ is the set of goal images. This is due to the fact that for every image that is not in the target set, we can always find a small particle to push towards the target set and decrease the value of the Lyapunov function. Given this Lyapunov function in \eqref{eq:lyapunov}, we choose a greedy feedback policy that minimizes the Lyapunov function from its current value at every given timestep. 

\begin{equation}
    u^*=\text{arg\,min}_u \bar{V}(\bar{f}(\mathbf{I}_k,u)).
    \label{eq:lyapunovoptimization}
\end{equation}

The intuitive explanation of this controller is to enforce the closed-loop behavior of the system to resemble a bowl with the target set as the flat bottom region (see Fig. \ref{fig:lyapunov}). Deforming the cutting board into a bowl, the carrot pieces will naturally fall on the target set. We hypothesize that depending on the accuracy of the prediction model $\mathbf{I}_{k+1}=\bar{f}(\mathbf{I}_k,u)$, the ability to descend along the Lyapunov function will differ. A better model should be able to descend faster along the Lyapunov function, and we use this descent curve as a task-relevant benchmark of the predictive capability of different models.

Furthermore, we note that if the set $\mathcal{S}_d$ is non-convex in Cartesian coordinates of $\mathbb{R}^2$, then $V(\mathcal{X})$ becomes non-convex as well. However, in image space, $V(y_k)$ is still linear (thus globally convex) function if $\|\mathbf{I}_{k}\|_{1,1}$ is relatively constant for all $k$ \eqref{eq:lyapunov}. This process is similar to convex relaxations, where originally non-convex problems are convexified in the space of distributions. Thus, a simple greedy strategy will stabilize to even non-convex target sets. Finally, we exclude sets that cannot be stabilized due to inherent mechanical limitation of the system, such as the width of the pusher.

\section{Models for Image Prediction}
\subsection{Switched-Linear Model}
\subsubsection{Model Description.}
In this model, we assume a switched-linear system \cite{switchedsystem}, with the following form:
\begin{equation}
    y_{k+1}=\mathbf{A}_i y_k,
    \label{eq:linear}
\end{equation}
where $\mathbf{A}_i\in\mathbb{R}^{N^2\times N^2}$ and $i=\{1,\cdots,|\mathcal{U}|\}$ with $\mathcal{U}$ the discretized action space. The action is directly represented as choosing the $\mathbf{A}$ matrix. One of the challenges of working in image space is that the coordinates of the actual object is recorded in the indices of the pixels, not the actual pixel values. Then, for a given action, such a linear map $\mathbf{A}$ can act as a permutation matrix that transports pixels in indices.

\subsubsection{Learning of Dynamics via Least Squares.}\label{lsqdyn}
To train the model, we collect many pairs of $(y_{k+1},y_k)$ that are all \textit{subject to the same action}. Let there be $M$ such pairs. Then, we construct a data matrix by appending the column vectors horizontally. Thus, we have two matrices $\mathbf{Y}_{k+1},\mathbf{Y}_k\in\mathbb{R}^{N^2\times M}$, where the $i^{th}$ column of $\mathbf{Y}_{k}$ is the vectorized image before the push, and the $i^{th}$ column of $\mathbf{Y}_{k+1}$ is the corresponding vectorized image after the push. Then, optimal identification of the transition matrix $\mathbf{A}$ can be written as a solution to the least squares optimization problem:
\begin{argmini}
   {\mathbf{A}}{\|\mathbf{Y}_{k+1}-\mathbf{A}\mathbf{Y}_k\|_F,}{}{\mathbf{A}^*=}
\end{argmini}
where $\|\cdot\|_F$ denotes the Frobenius norm. This is a matrix Ordinary Least Squares (OLS) problem, and has a standard closed-form solution.

We attempt to also add constraints or add regularization terms and quantify which least-squares method performs the best. We questioned if imposing a permutation matrix-like structure could provide meaningful regularization for the $\mathbf{A}$ matrix, and improve the test performance of prediction. We also considered non-negativity constraints, or equality constraints such that the rows or columns sum up to $1$. These optimization problems can be easily formulated as an instance of Quadratic Programming (QP). The general formulation of these constraints is written down in \eqref{eq:qp}.
\begin{argmini}
    {\mathbf{A}}{\|\mathbf{Y}_{k+1}-\mathbf{A}\mathbf{Y}_k\|_F}{}{\mathbf{A}^*=}
    \addConstraint{\mathbf{A}\geq 0 ,\quad\textstyle\sum_i \mathbf{A}[i,j]=1 \quad \forall j.}
    \label{eq:qp}
\end{argmini}
While the transcription to QP is trivial, the number of decision variables in $\mathbf{A}\in\mathbb{R}^{N^2\times N^2}$ is $N^4$, which can reach a million even for a $32\times 32$ resolution image. Thus, the full-scale problem is very time-consuming to solve. However, for row-constrained or non-negative constraints, we can utilize the following relation between the row of $\mathbf{Y}_{k+1}$ and the row of $\mathbf{A}$:
\begin{equation}
    \mathbf{Y}_{k+1}[i,:] = \mathbf{A}[i,:]\mathbf{Y}_k,
    \label{eq:submodularity}
\end{equation}
where $\mathbf{A}[i,:]$ denotes the $i^{th}$ row of $\mathbf{A}$. Using this relation, it is possible to decompose this optimization problem into $N^2$ optimization problems with $N^2$ decision variables, since the optimal solution of $\mathbf{A}[i,:]$  is the optimal solution for the entire problem. 

While the column-sum-constrained problem also presents an interesting interpretation as a Markov stochastic matrix \cite{stochasticmatrix}, it no longer possesses this natural decomposition, and the resulting problem is too big to handle for regular QP solvers, requiring more scalable formulations such as ADMM \cite{jack1,admm}. We plan to explore more methods of dealing with large-scale constrained least squares in future works. 

\subsubsection{Tensor Description of the Transition Matrix.}

\begin{figure}[b]
\centering
\includegraphics[height=3.3cm]{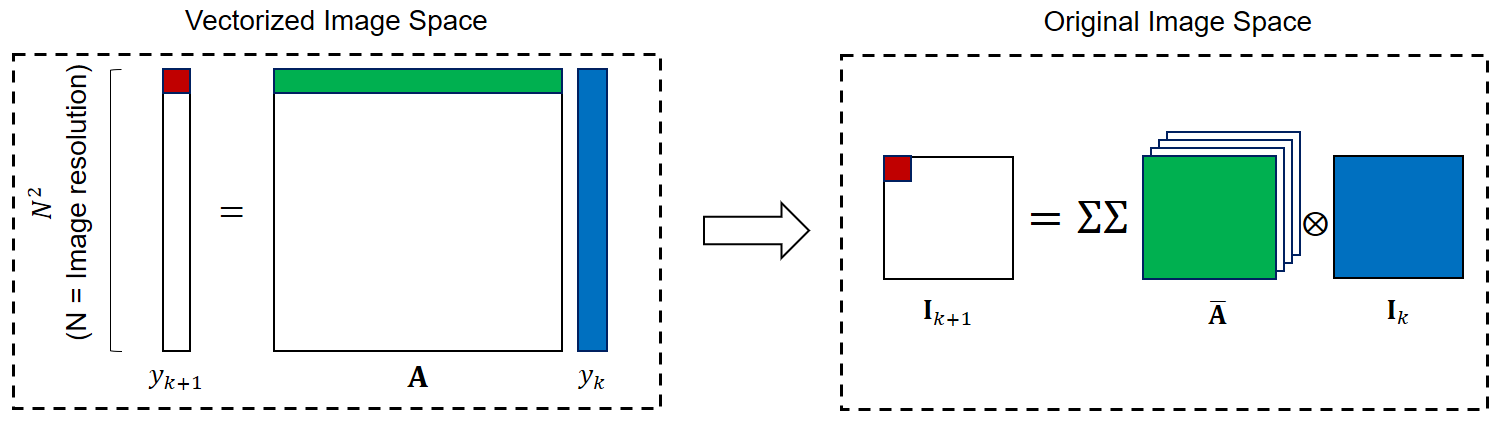}
\caption{In the vectorized image space, each element (red) is the result of a dot product between the corresponding row of the $\mathbf{A}$ matrix (green) and the original vector (blue). By refactoring the green row vector as an image, we can represent the red pixel as a sum of element-wise multiplication between the green image and the blue image.}
\label{fig:tensor}
\end{figure}

To interpret the $\mathbf{A}$ matrix, let us reshape $\mathbf{A}\in\mathbb{R}^{N^2\times N^2}$ into a tensor $\mathbf{\bar{A}}\in\mathbb{R}^{N\times N \times N \times N}$. Similar to how $\mathbf{A}[i,j]$ denotes how much the $j^{th}$ element of $y_k$ affects the $i^{th}$ element of $y_{k+1}$, $\mathbf{\bar{A}}[i,j,m,n]$ signifies how much the pixel value of $\mathbf{I}_k[m,n]$ affects the pixel value of $\mathbf{I}_{k+1}[i,j]$. Using this relation, we can rewrite \eqref{eq:linear} as
\begin{equation}
    \mathbf{I}_{k+1}=\mathbf{\bar{A}}\times_2 \mathbf{I}_k,
\end{equation}
where $\times_2$ denotes the $2$-mode tensor product over the last two dimensions.

When the matrix is reshaped as a tensor, the structure of the tensor reveals an interesting fact about the linear model. The rows of the $\mathbf{A}$ matrix can be reshaped as an $N\times N$ image, as $\mathbf{A}[i,:]$ corresponds to $\mathbf{\bar{A}}[i,j,:,:]$ in the tensor representation. Then, the matrix $\mathbf{\bar{A}}[i,j,:,:]$ gets element-wise multiplied with the original image, and the grand sum of this product matrix becomes the value of the pixel in $\mathbf{I}_k[i,j]$. We call the image $\mathbf{\bar{A}}[i,j,:,:]$ the kernel of the pixel $\mathbf{I}_k[i,j]$, and use it as a convenient visualization of the learned linear model in the results section (Fig. \ref{fig:kernelvisualization}).

\subsubsection{Affine Transformation on Input Images.}

\begin{figure}[t]
\centering
\includegraphics[height=2.1cm]{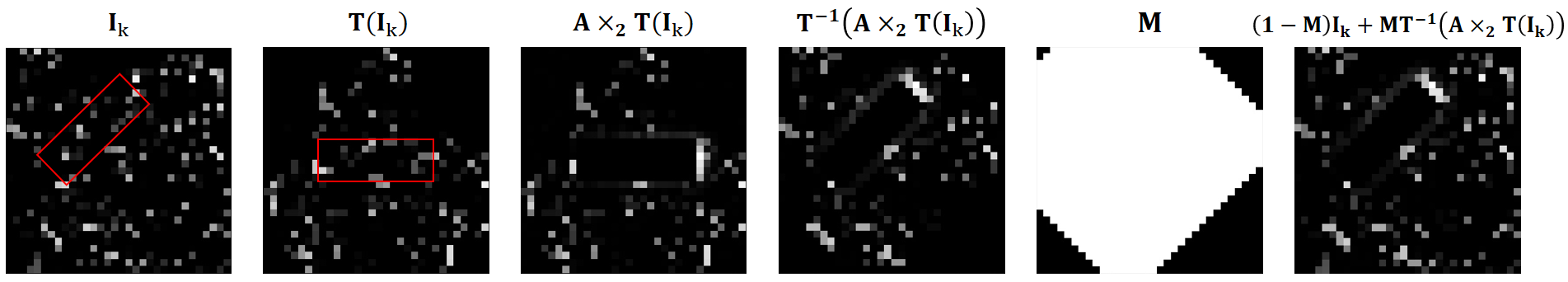}
\caption{Affine Transformation Process for Reducing Action Space Dimension}
\label{fig:affinetransform}
\end{figure}

The above process explains how to find the optimal $\mathbf{A}$ for a given action $u$, but we need to learn every $\mathbf{A}$ for all the different inputs. We note that our action space is in a Cartesian frame, and it should be possible to utilize coordinate transforms to reduce the dimension of the action space, similar to the approach taken in \cite{affinetransformation}.

Given an action $u$, we can create a rectangle that represents the area swept by the pusher, which we call the ``push rectangle''. Then, we compute $\mathbf{T}\in SE(2)$, an affine transformation from the center-of-image frame to pusher rectangle frame. Let $\mathbf{T}(\mathbf{I})$ be the transformed image. We apply our linear map to the transformed image, and transform the predicted image back to original coordinates by the inverse transform $\mathbf{T}^{-1}$. To deal with the missing parts of images in the transformation process, we let a mask go through the same affine transformation $\mathbf{M}=\mathbf{T}^{-1}\left(\mathbf{T}\left(\mathbf{1}^{N\times N}\right)\right)$, which is used to combine the predicted image and the original image. This process is illustrated in Fig. \ref{fig:affinetransform}. 

With the affine transforms on input, we only need to compute different $\mathbf{A}_i$ matrices for the length of the push, which significantly decreases the amount of data we need to have. We discretize push length into $5$, and train an $\mathbf{A}$ matrix for each one of them with $1000$ sample pairs of $(\mathbf{I}_k,\mathbf{I}_{k+1})$.

\subsection{Deep-Learning Model}

\begin{figure}[t]
\centering
\includegraphics[height=3.4cm]{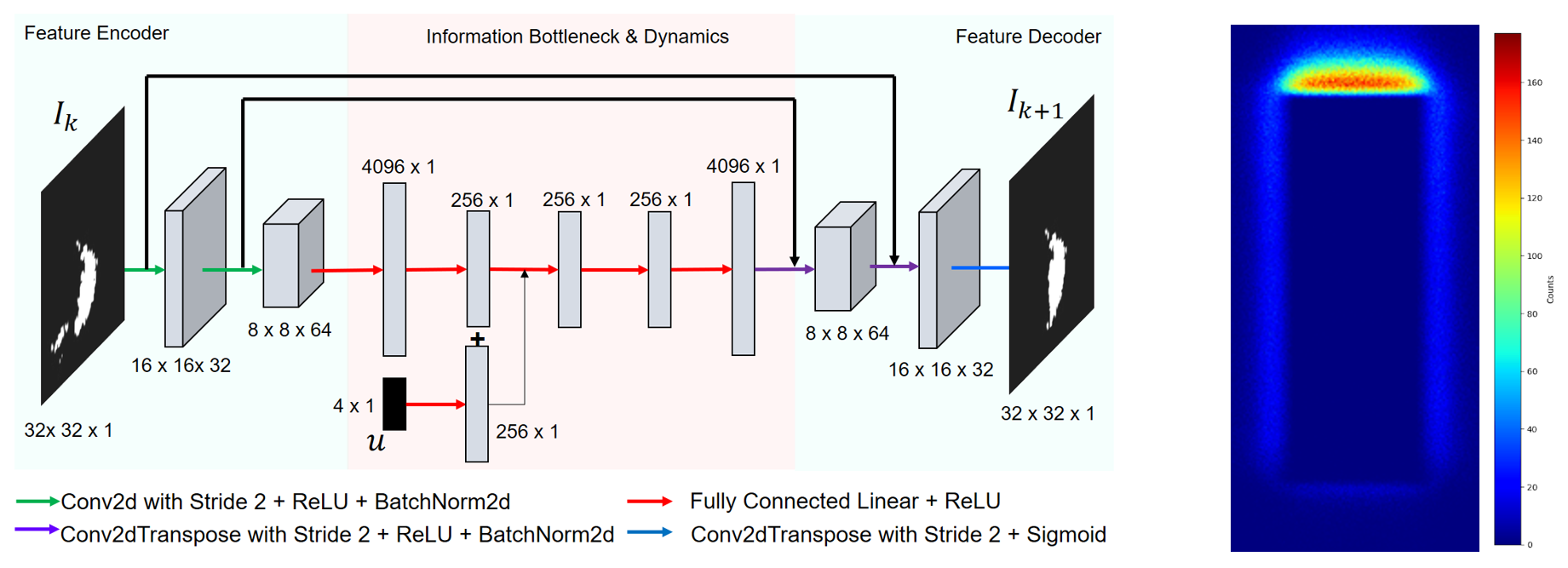}
\caption{Left: deep-learning Architecture for the DVF-Original Model. Right: Learned distribution of the object-centric model.}
\label{fig:objectoriented}
\vskip -0.2cm
\end{figure}

We use a deep-learning model based on \cite{Finn1,dvf2,prom,spatialautoencoder}, which was our original approach to this problem. This architecture (Fig.\ref{fig:objectoriented}) computes the latent-space vectors of the image $\mathbf{I}_k$ through a convolutional autoencoder \cite{cae}, approximates the dynamics on this latent-space vector with multiple layers of Multi-Layer Perceptrons (MLP), then decodes the resulting latent-space vector to obtain the predicted image $\mathbf{I}_{k+1}$. In addition, similar to \cite{Finn1,dvf2,prom}, skip connections between matching dimensions of the convolutional autoencoder are added to preserve high-frequency features that are lost during convolution. This architecture is shown in Fig.\ref{fig:objectoriented}, and we will label it \textit{DVF-Original}. The network is trained with $23,000$ pairs of $(\mathbf{I}_k,u,\mathbf{I}_{k+1})$. 

Although majority of similar architectures append the action with the latent-space states and compute the dynamics, we wanted to make a fair comparison with the linear model which utilizes affine transforms on the input image, such that the network is also learning image-to-image translation instead of image$\times$action-to-image. Thus, we utilize the same method illustrated in Fig.\ref{fig:affinetransform}, and use $5$ separate networks for different push lengths, similar to the mixture-of-experts approach. The action-appending part of the network is removed in order to facilitate an image-to-image architecture. We call this method \textit{DVF-Affine}. 

Let the forward predicting function be $\mathbf{\hat{I}}_{k+1}=\bar{f}(\mathbf{I}_k,u)$. Each network is trained end-to-end in a supervised manner using the true outcome image, and the loss function is defined as:
\begin{equation}
    \mathcal{L}(\mathbf{\hat{I}}_{k+1},\mathbf{I}_{k+1})=\|\mathbf{\hat{I}}_{k+1}-\mathbf{I}_{k+1}\|_F=\|\bar{f}(\mathbf{I}_k,u)-\mathbf{I}_{k+1}\|_F.
\end{equation}
The samples that were used for training the linear model is used to train each network in DVF-Affine. The training was done using the ADAM optimizer \cite{adam}, and each network went through more than $1000$ epochs. The first $500$ epochs are trained with a step learning rate scheduler, starting with $\eta=0.01$ and decreasing by a factor of $10$ every $100$ epochs ($\gamma=0.1$). This process is repeated again for another $500$ epochs, resuming from the result of the first $500$ epochs.  

\subsection{Object-Centric Transport Model}\label{section:objectoriented}
Although the true states of the objects are unobservable, we attempt to build a first-principles model by treating each non-zero pixel in the image as an object. We assume that pixels that are within the swept area of the push are teleported by a probability distribution around the pusher. The algorithm works as follows:
\begin{enumerate}\itemsep-0.1em
    \item From image $\mathbf{I}_k$, collect the coordinates of the non-zero pixels as a set $\mathcal{X}_k$
    \item Divide $\mathcal{X}_k$ into $\mathcal{X}_{a}$ (affected) and $\mathcal{X}_{u}$ (unaffected) depending on whether or not the pixel coordinate is within the push rectangle. 
    \item From a bivariate distribution $P(u)$, sample $|\mathcal{X}_a|$ coordinates $p_i\sim P$, and denote the new set of sampled coordinates as $\mathcal{X}_{n}$. 
    \item Create a new set $\mathcal{X}_{k+1}=\mathcal{X}_n \cup \mathcal{X}_u$
    \item Evaluate the particle Lyapunov function in \eqref{eq:lyapunovparticle} directly.
    \vskip -0.15 true in
\end{enumerate}

We approximate the distribution $P(u)$ by a small uniform distribution right in front of the push rectangle. To justify this choice, we transform each image using the affine transformation from Fig. \ref{fig:affinetransform}, evaluate the difference between the image ($\mathbf{I}_{k+1}-\mathbf{I}_k$), then threshold the difference above zero. This distribution is illustrated in the right side of Fig. \ref{fig:objectoriented}, and we can see that the uniform distribution approximates this distribution well.

\section{Simulation Results}
\begin{figure}[t]
\centering
\includegraphics[height=2.8cm]{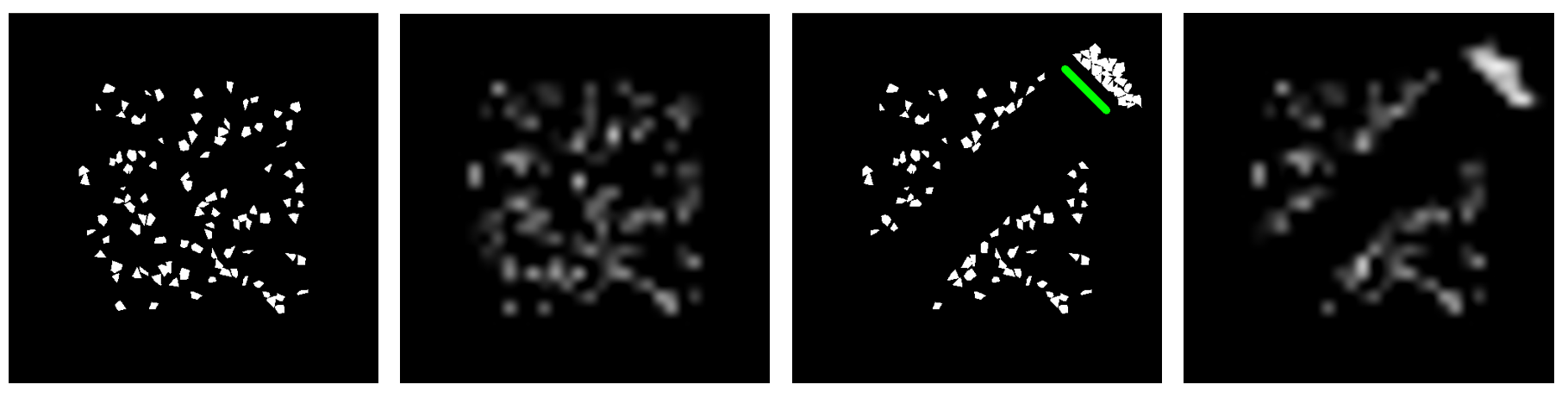}
\caption{The left two images visualize the simulator environment at time $k$, and the downsampled image $\mathbf{I}_k$ that acts as input to the prediction algorithm. The right two images corresponds to time $\mathbf{I}_{k+1}$ after the push.}
\label{fig:simulator}
\end{figure}

To learn prediction models and test the closed-loop performance, we built a simulator using Pymunk. Each carrot piece is randomly generated by sampling points along a fixed-size circle and computing their convex hull. The simulator is displayed in Fig. \ref{fig:simulator}. 

\subsection{Switched-Linear Model}
\subsubsection{Least-Squares Comparison Results.}

\begin{figure}[b]
\centering
\includegraphics[height=2.3cm]{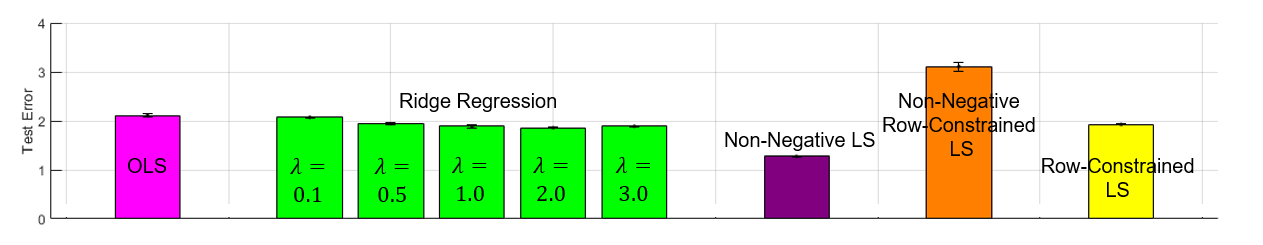}
\caption{Test error on different least-squares algorithms}
\label{fig:lsqcomparison}
\end{figure}

We compare different methods for least-squares from Sec. \ref{lsqdyn}. Around $1000$ pairs of images ($\mathbf{I}_k,\mathbf{I}_{k+1}$) are collected, where each image is $32\times 32$. $800$ pairs are used to estimate the optimal value of $\mathbf{A}$ using the CVXOPT solver \cite{cvxopt}, while $200$ pairs are used as test set. This result is shown in Fig. \ref{fig:lsqcomparison}, and the non-negative least-squares formulation worked best for the estimation of the linear map. On the other hand, the row sum-constrained least-squares solution seem to regularize the transition matrix excessively, as the sum of all kernels are forced to be equal to $1$. Therefore, we adopt non-negative least squares as the default estimation scheme for the $\mathbf{A}$ matrix.
\subsubsection{Learned Kernels of the Linear Model.}
\begin{figure}[t]
\centering
\includegraphics[height=4.20cm]{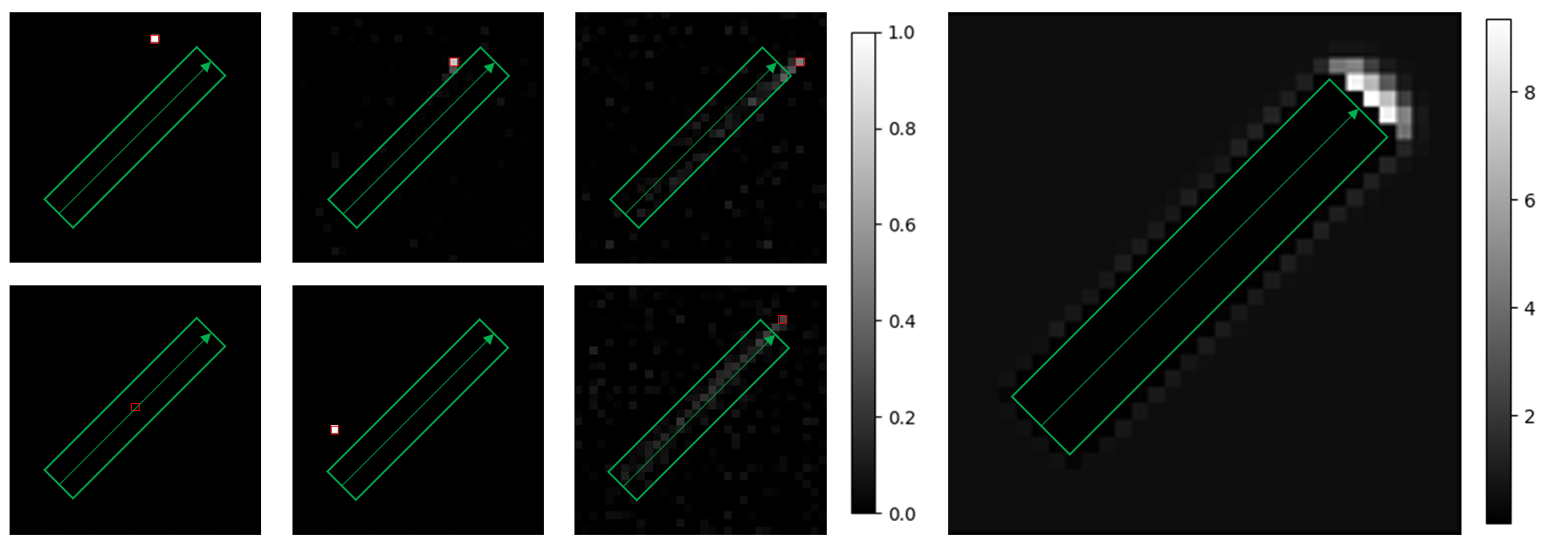}
\caption{Visualization of different kernels for different pixel locations. The red pixel represents the location of the pixel of the kernel, and the green rectangle represents the pushed area. The right image represents the ``step response'' of the matrix.}
\label{fig:kernelvisualization}
\vskip -0.5cm
\end{figure}

Following our interpretation of the $\mathbf{A}$ matrix as a tensor in Fig. \ref{fig:tensor}, we visualize each of the kernels to see if the kernel images are interpretable. The result of this visualization is shown in Fig.  \ref{fig:kernelvisualization}

We see that for the areas outside the push rectangle, the kernel learns the identity transform. For areas inside the push rectangle, the kernel values are almost zero, as it learns that pixels will be gone from this location. Finally, for pixels that are at the edge of the push rectangle, it learns a kernel to weight the values inside the pushed area and sum them up to place them in front of the pushed area.

In addition, we initialize $y_k$ to all $0.5$ (half of maximum value), and see the step response of $y_{k+1}=\mathbf{A}y_k$ to see the behavior of the $\mathbf{A}$ matrix. The result is displayed in the right side of Fig. \ref{fig:kernelvisualization}, and we see that the transition matrix learns the correct behavior of the action $u$. It is surprising to see the similarity between this step response and the probability distribution obtained in the transport model (Fig.  \ref{fig:objectoriented}), given that the linear model is obtained entirely from data while the transport model is relatively hand-written.
\vskip -0.5cm
\subsection{Comparison of Visual Prediction} 

\begin{figure}[h]
\centering
\includegraphics[height=2.65cm]{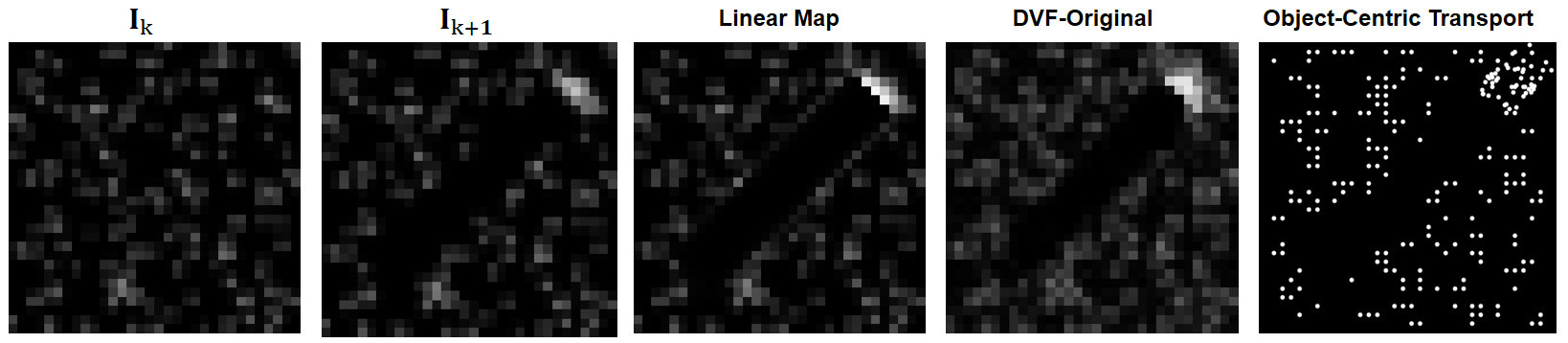}
\caption{Visual comparison of Model Prediction Results.}
\label{fig:visualcomparison}
\end{figure}

We compare the results of visual prediction using the three models. For fairness, all models utilize a $32\times 32$ resolution image. The predictions are visualized in Fig. \ref{fig:visualcomparison}. We observe that both the linear model and deep models predict a reasonable result of the push. The Object-Centric model is visualized with locations of particles as it does not synthesize future image frames. 

In order to quantify performance, we set up $1000$ test examples generated from different initial conditions and actions. We take $\|\mathbf{I}_{k+1}-f(\mathbf{I}_k,u)\|_F$ as the error metric, and average the error across all $1000$ samples. This test error is illustrated in Table. \ref{tab:predictionerror}. Although the deep models were trained for over $1000$ epochs, the linear model has the lowest prediction error. Given that the linear model is a subclass of the deep model and has less parameters compared to the deep model, the fact that it performs better is surprising. 

We believe this is an empirical evidence for some underlying linearity in the problem, and thus the linear model has better inductive bias. Furthermore, choosing a linear model allowed us to train the model \eqref{eq:qp} to global optimality, and add meaningful regularization constraints such as non-negativity.  
\vskip -0.5cm
\begin{table}[]
\caption{Test Prediction Error, Model Details, and Evaluation Time}
\centering
\begin{tabular}{|l|l|c|c|c|r|}
\hline 
Model Name  & Dimension & Parameters & Samples & Test Error \\\hline 
Switched-Linear & $\mathbb{R}^{N\times N}\rightarrow \mathbb{R}^{N\times N}$   & 1,048,576  & 5,000  & 1.858  \\
DVF-Original    & $\mathbb{R}^{N\times N}\times \mathbb{R}^4\rightarrow \mathbb{R}^{N\times N}$ & 2,382,721 & 23,000 & 2.062     \\
DVF-Affine      & $\mathbb{R}^{N\times N}\rightarrow \mathbb{R}^{N\times N}$ & 2,317,185 & 5,000 & 2.537  \\ 

\hline 
\end{tabular}
\label{tab:predictionerror}
\vskip -1cm
\end{table}

\subsection{Closed-Loop Performance}

\begin{figure}[b!]
\vskip -0.2cm
\centering
\includegraphics[height=5.7cm]{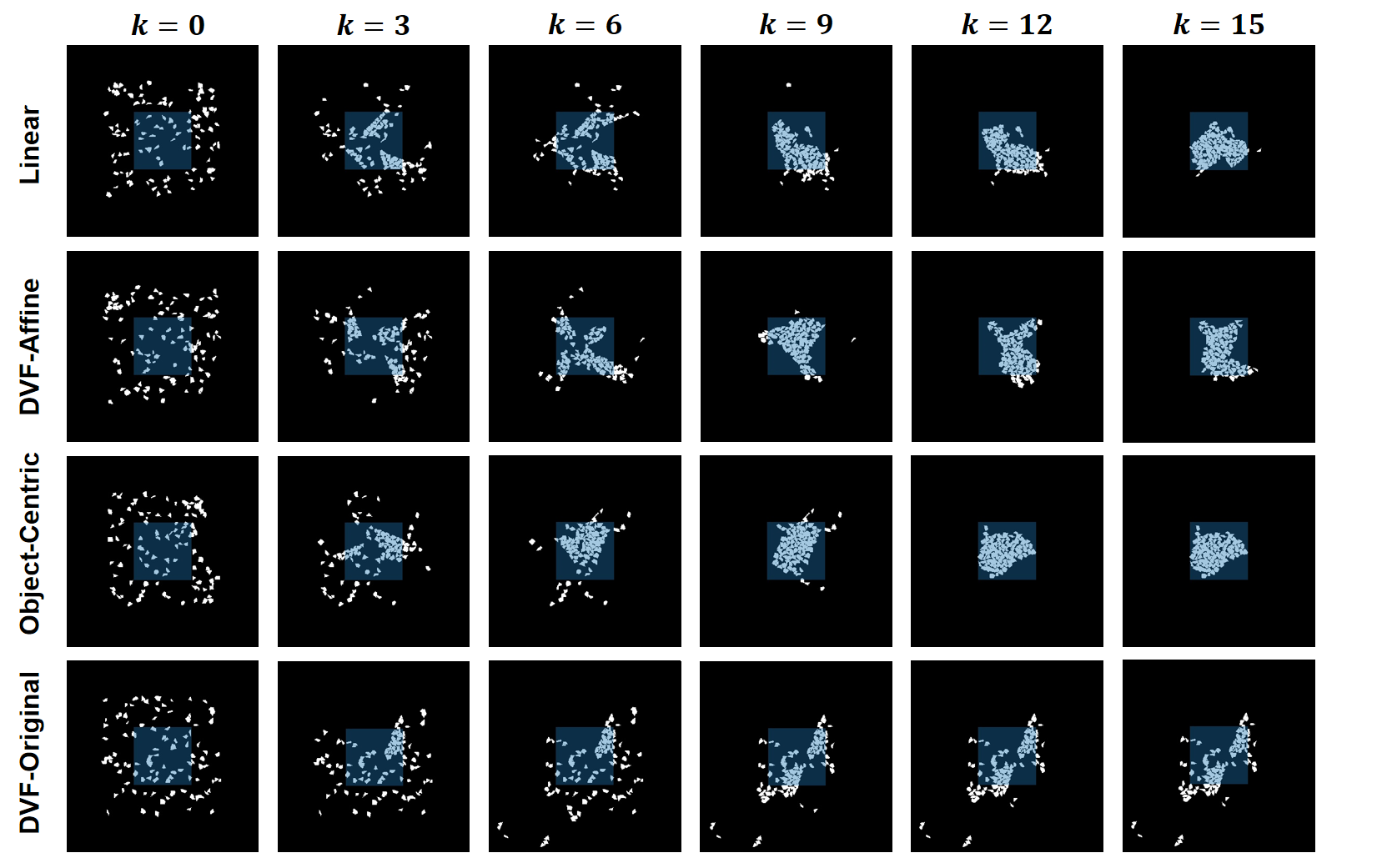}
\caption{Visualization of the closed-loop behavior using three predicted models. The blue square region denotes the target set.}
\label{fig:closedloopvisualization}
\end{figure}

\begin{figure}[t!]
\centering
\includegraphics[width=0.85\linewidth]{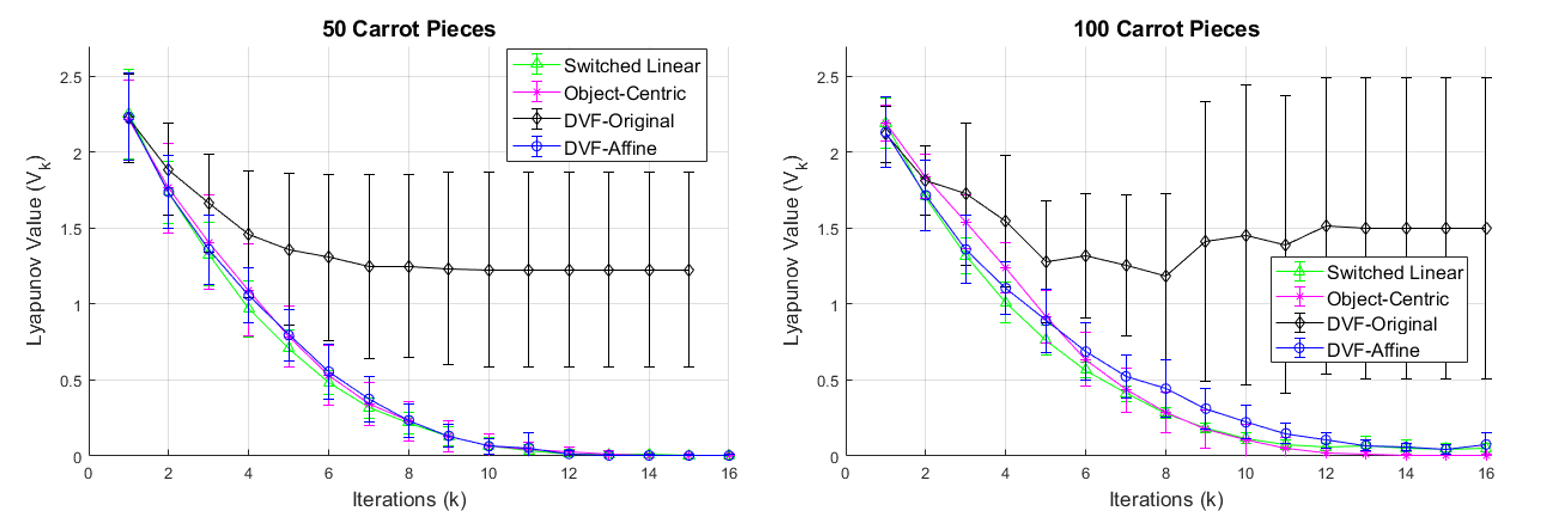}
\caption{Evaluation of descent along the Lyapunov function for different methods. The goal of the task is to make $V=0$. Each method was repeated for $10$ times on different initial conditions.}
\label{fig:closedloopresult}
\vskip -0.3cm
\end{figure}

We compare the performance of the models in closed-loop by evaluating the best action on a fixed grid of $u$ according to \eqref{eq:lyapunovoptimization}. The goal of the task is to push all the pieces into the blue region and drive the Lyapunov value $V$ to $0$. Some image trajectories are visualized in Fig. \ref{fig:closedloopvisualization}, and the result is plotted in Fig. \ref{fig:closedloopresult}.

We observed that while the linear model, the object-centric transport model, and DVF-Affine model are able to converge to the target set, DVF-Original failed to do so. A common failure mode in the DVF-Original model is that its optimal predicted action does not cause any change in the actual scene, and the same image gets feedbacked, making the closed-loop system get stuck in a loop. This failure mode agrees with the observation in \cite{Tucker}, where policies trained in the latent-space of spatial autoencoders \cite{spatialautoencoder} repeatedly produced actions that did not change the scene. We observed these failure cases and found that deep networks mispredict some key physical behavior that is apparent in object pushing, such as making carrots disappear instead of pushing them. 

There was no significant performance difference between the remaining three models with small number of carrots. However, for larger number of carrot pieces, the linear model started showing a steeper descent curve compared to the DVF-Affine and object-centric models, with DVF-Affine usually requiring $2$ more iterations before convergence. On average, the switched-linear model took 1.00 second of computation, objected-oriented took 27.00s, and DVF-affine model took 4.00s on the CPU, while the DVF-original took 0.17s on GPU.

\begin{figure}[b!]
\centering
\includegraphics[height=4.65cm]{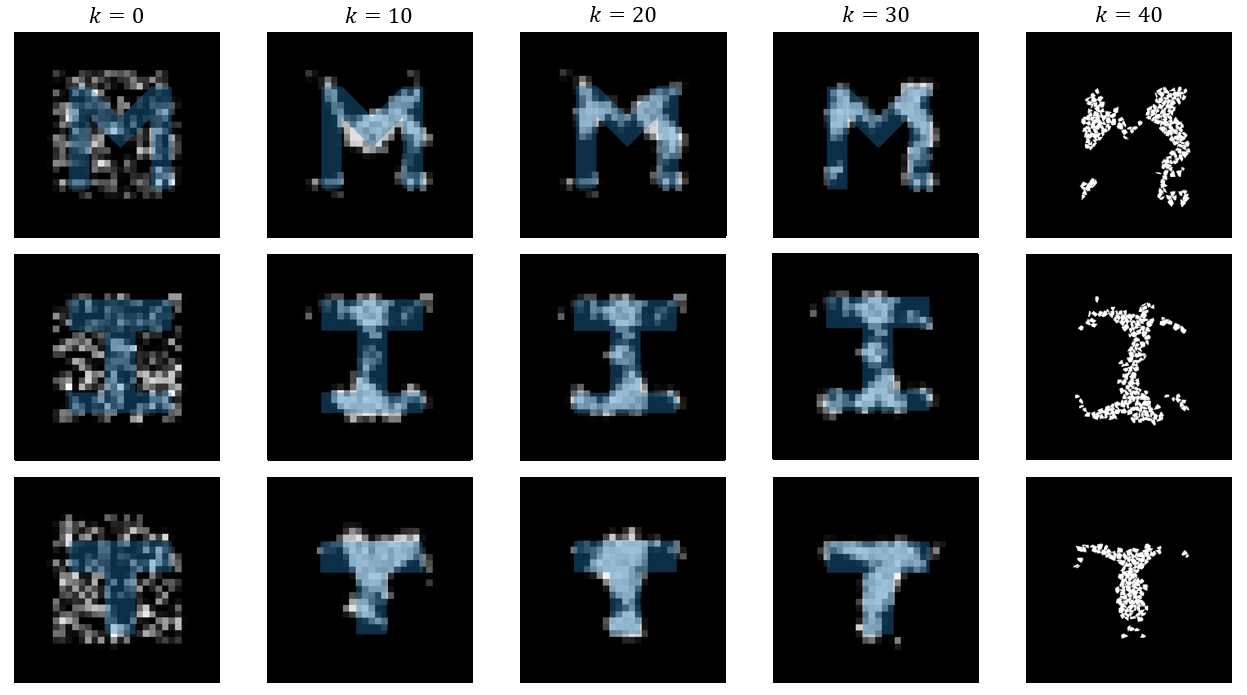}
\caption{Evaluation of linear model on more difficult target sets. The last images are in original simulator resolution.}
\label{fig:nonconvex}
\end{figure}

We attempt more difficult target sets to showcase the ability of the linear model, as shown in Fig. \ref{fig:nonconvex}. Although it takes more iterations, the prediction of the linear model is successful in converging to a complex non-convex target set when coupled with the controller. 

\section{Experiment Results}
\vskip -0.2cm
\begin{figure}[t]
\centering
\includegraphics[width=0.87\linewidth]{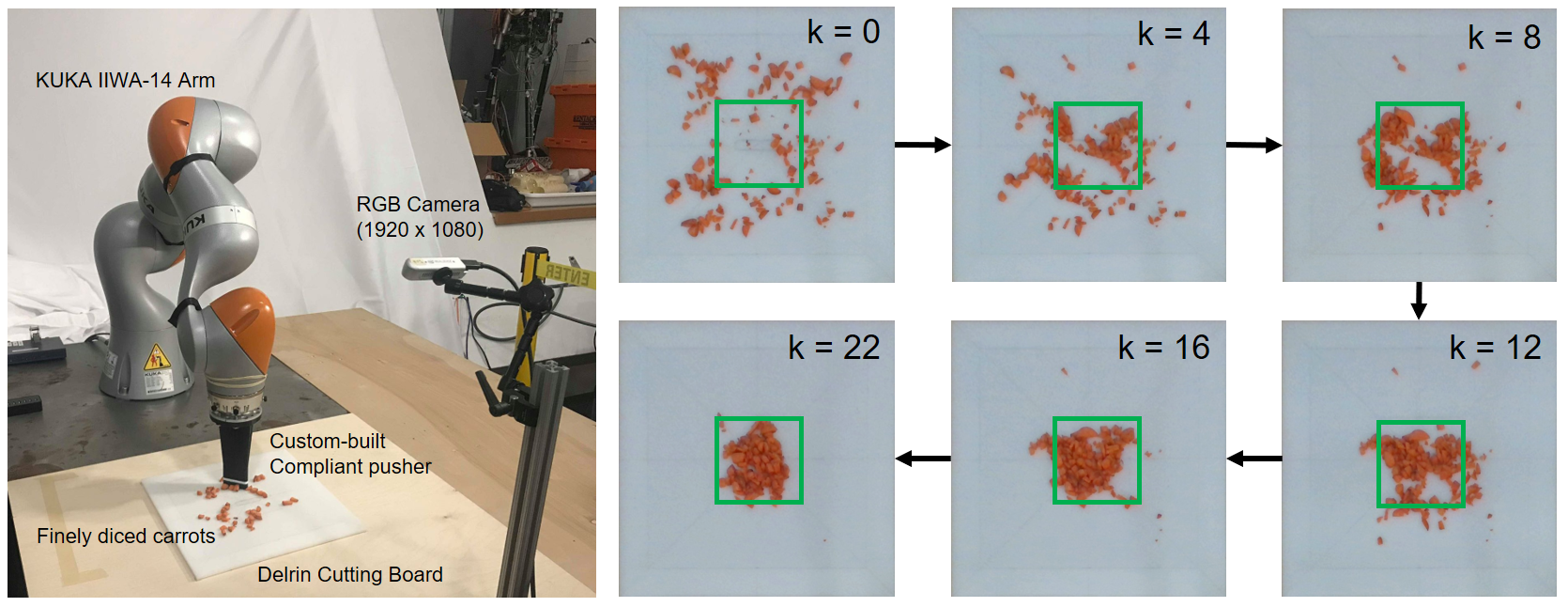}
\caption{Left: Experiment of the setup. Right: Visualization of the closed-loop behavior of the linear model.}
\label{fig:experimentsetup}
\vskip -0.3cm
\end{figure}

To test our algorithm in the real-world, we prepare an experiment setup illustrated in Fig. \ref{fig:experimentsetup}. The initial piles are generated by manually spraying the carrot pieces on the board, and the result is averaged over multiple runs. We use the dynamics obtained in simulation directly on the experiment without additional fine-tuning, as we believe this sim-to-real transfer process will be a good measure of the model's ability to generalize. 

\begin{figure}[b!]
\centering
\includegraphics[height=4.3cm]{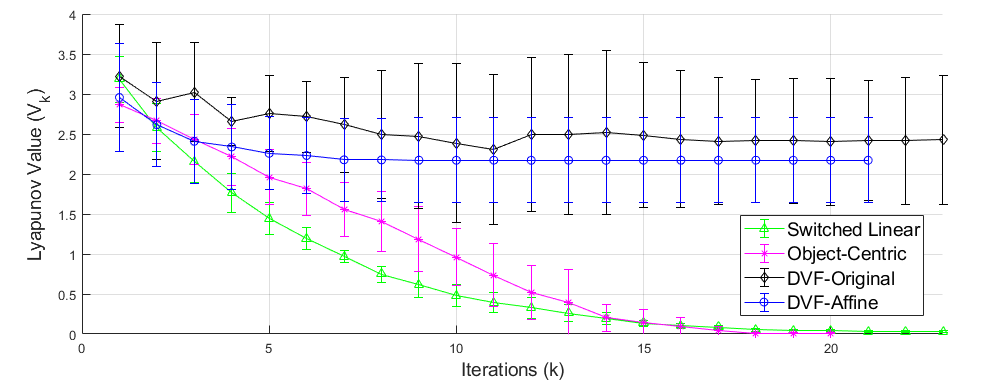}
\caption{Result of the experiments. Each method was repeated 5 times.}
\label{fig:closedloopresultreal}
\vskip-0.2cm
\end{figure}

From the result plot of Fig. \ref{fig:closedloopresultreal}, we see that the linear model and the object-centric model was still able to converge to the target set. But surprisingly, the \textit{DVF-Affine model did not succeed} in the real-world task, despite its success in simulation. This suggests that the DVF-Affine model overfitted to the training images provided by the simulator, while the switched-linear model learned a generalizable model that can extend to new environments.

What could have allowed the linear model to generalize to the real environment while the DVF-Affine model did not? Given that they are trained with the sample training samples, we believe that the linear model provided better inductive bias for the phenomenon, which empirically signifies an inherent linearity in the problem. 

\vskip-0.2cm
\section{Conclusion}
\vskip -0.2cm
In this work, we have proposed a switched-linear model that utilizes action-dependent linear maps to predict image dynamics. We compared the performance of our model with deep-learning-based models which estimate the dynamics on the latent space of images. We found that that the linear model outperformed the deep model in test prediction error, closed-loop performance, and generalization. Furthermore, through comparing the linear model's performance with the object-centric model, we found that output feedback offers a competitive alternative to our object-centric first-principles approach.

As linear models are a subclass of deep models, we believe that given enough training samples, the right architecture, and good hyperparameters, there exists a deep model that will outperform the linear model. However, the search procedure for such models is not understood well. On the other hand, linear models allow globally optimal learning in their parameter space, and can easily be regularized through constraints. Furthermore, this relatively simple task serves to illustrate that high-dimensional visual dynamics, that may not seem linear at first glance, show promise to be approximated well by tractable linear models.

We wish to better investigate where this linearity comes from, and how general this approach will be for manipulation tasks such as pushing rigid bodies, tasks in 3D, and tasks that require utilizing the color-space. We aim to understand what model classes are most useful for vision-based feedback in manipulation, considering both the quality of visual prediction, as well as compatibility with rigorous methods for system identification, control design, and analysis.
\vskip -0.5cm
%
\bibliography{mybib}

\begin{thebibliography}{10}

\bibitem{pulkit1}
Agrawal, P., Nair, A., Abbeel, P., Malik, J., Levine, S.:
\newblock Learning to poke by poking: Experiential learning of intuitive
  physics.
\newblock CoRR \textbf{abs/1606.07419} (2016)

\bibitem{cvxopt}
Andersen, M., Dahl, J., Vandenberghe, L.:
\newblock Cvxopt: A python package for convex optimization, version 1.1.6
  (2013)

\bibitem{chamfer}
Barrow, H.G., Tenenbaum, J.M., Bolles, R.C., Wolf, H.C.:
\newblock Parametric correspondence and chamfer matching: Two new techniques
  for image matching.
\newblock In: Proceedings of the 5th International Joint Conference on
  Artificial Intelligence - Volume 2, Morgan Kaufmann Publishers Inc. (1977)
  659–663

\bibitem{admm}
Boyd, S., Parikh, N., Chu, E., Peleato, B., Eckstein, J.:
\newblock Distributed optimization and statistical learning via the alternating
  direction method of multipliers.
\newblock Found. Trends Mach. Learn. \textbf{3}(1) (January 2011)  1–122

\bibitem{switchedsystem}
Colaneri, P.:
\newblock Analysis and control of linear switched systems.
\newblock Politecnico di Milano (2015)

\bibitem{8460915}
{Elliott}, S., {Cakmak}, M.:
\newblock Robotic cleaning through dirt rearrangement planning with learned
  transition models.
\newblock In: 2018 IEEE International Conference on Robotics and Automation
  (ICRA). (2018)  1623--1630

\bibitem{dvf2}
Finn, C., Goodfellow, I., Levine, S.:
\newblock Unsupervised learning for physical interaction through video
  prediction.
\newblock In: Proceedings of the 30th International Conference on Neural
  Information Processing Systems, Red Hook, NY, USA (2016)  64–72

\bibitem{Finn1}
Finn, C., Levine, S.:
\newblock Deep visual foresight for planning robot motion.
\newblock CoRR \textbf{abs/1610.00696} (2016)

\bibitem{spatialautoencoder}
Finn, C., Tan, X.Y., Duan, Y., Darrell, T., Levine, S., Abbeel, P.:
\newblock Learning visual feature spaces for robotic manipulation with deep
  spatial autoencoders.
\newblock CoRR \textbf{abs/1509.06113} (2015)

\bibitem{billiard}
Fragkiadaki, K., Agrawal, P., Levine, S., Malik, J.:
\newblock Learning visual predictive models of physics for playing billiards.
\newblock CoRR \textbf{abs/1511.07404} (2015)

\bibitem{stochasticmatrix}
Gagniuc, P.:
\newblock Markov Chains: From Theory to Implementation and Experimentation.
\newblock Wiley (2017)

\bibitem{difftaichi}
Hu, Y., Anderson, L., Li, T., Q, S., Carr, N., Ragan-Kelley, J., Durand, F.:
\newblock Difftaichi: Differentiable programming for physical simulation.
\newblock CoRR \textbf{abs/1412.6980} (2014)

\bibitem{adam}
Kingma, D.P., Ba, J.:
\newblock Adam: A method for stochastic optimization.
\newblock CoRR \textbf{abs/1412.6980} (2014)

\bibitem{cnn}
Krizhevsky, A., Sutskever, I., Hinton, G.E.:
\newblock Imagenet classification with deep convolutional neural networks.
\newblock Commun. ACM \textbf{60}(6) (May 2017)  84–90

\bibitem{lasserre_nonlinear_2008}
Lasserre, J.B., Henrion, D., Prieur, C., Trélat, E.:
\newblock Nonlinear {Optimal} {Control} via {Occupation} {Measures} and
  {LMI}-{Relaxations}.
\newblock SIAM Journal on Control and Optimization \textbf{47}(4) (January
  2008)  1643--1666

\bibitem{rodriguez2}
{Ma}, D., {Rodriguez}, A.:
\newblock Friction variability in planar pushing data: Anisotropic friction and
  data-collection bias.
\newblock IEEE Robotics and Automation Letters \textbf{3}(4) (Oct 2018)
  3232--3239

\bibitem{cae}
Masci, J., Meier, U., Cire{\c{s}}an, D., Schmidhuber, J.:
\newblock Stacked convolutional auto-encoders for hierarchical feature
  extraction.
\newblock In Honkela, T., Duch, W., Girolami, M., Kaski, S., eds.: Artificial
  Neural Networks and Machine Learning, Berlin, Heidelberg, Springer Berlin
  Heidelberg (2011)  52--59

\bibitem{mason1}
Mason, M.T.:
\newblock Mechanics and planning of manipulator pushing operations.
\newblock The International Journal of Robotics Research \textbf{5}(3) (1986)
  53--71

\bibitem{keypoints}
Minderer, M., Sun, C., Villegas, R., Cole, F., Murphy, K., Lee, H.:
\newblock Unsupervised learning of object structure and dynamics from videos.
\newblock CoRR (2019)

\bibitem{keto}
Qin, Z., Fang, K., Zhu, Y., Li, F.F., Savarese, S.:
\newblock Keto: Learning keypoint representations for tool manipulation.
\newblock ArXiv \textbf{abs/1910.11977} (2019)

\bibitem{prom}
Sarkar, M., Pradhan, P., Ghose, D.:
\newblock Planning robot motion using deep visual prediction.
\newblock CoRR \textbf{abs/1906.10182} (2019)

\bibitem{jack1}
{Umenberger}, J., {Manchester}, I.R.:
\newblock Scalable identification of stable positive systems.
\newblock In: IEEE 55th Conference on Decision and Control (CDC). (Dec 2016)
  4630--4635

\bibitem{vin}
Watters, N., Tacchetti, A., Weber, T., Pascanu, R., Battaglia, P.W., Zoran, D.:
\newblock Visual interaction networks.
\newblock CoRR \textbf{abs/1706.01433} (2017)

\bibitem{williams_datadriven_2015}
Williams, M.O., Kevrekidis, I.G., Rowley, C.W.:
\newblock A {Data}–{Driven} {Approximation} of the {Koopman} {Operator}:
  {Extending} {Dynamic} {Mode} {Decomposition}.
\newblock Journal of Nonlinear Science \textbf{25}(6) (December 2015)
  1307--1346

\bibitem{Tucker}
Wilson, M., Hermans, T.:
\newblock Learning to manipulate object collections using grounded state
  representations.
\newblock Conference on Robot Learning (2019)

\bibitem{gupta1}
Ye, Y., Gandhi, D., Gupta, A., Tulsiani, S.:
\newblock Object-centric forward modeling for model predictive control.
\newblock ArXiv \textbf{abs/1910.03568} (2019)

\bibitem{Tsujio1}
{Yong Yu}, {Fukuda}, K., {Tsujio}, S.:
\newblock Estimation of mass and center of mass of graspless and shape-unknown
  object.
\newblock In: IEEE International Conference on Robotics and Automation.
  Volume~4. (May 1999)  2893--2898 vol.4

\bibitem{affinetransformation}
Zeng, A., Song, S., Welker, S., Lee, J., Rodriguez, A., Funkhouser, T.A.:
\newblock Learning synergies between pushing and grasping with self-supervised
  deep reinforcement learning.
\newblock CoRR \textbf{abs/1803.09956} (2018)

\end{thebibliography}
\bibliographystyle{splncs_srt}
\end{document}